\pdfoutput=1

\documentclass[11pt]{article}

\usepackage{naacl2021}

\usepackage{times}
\usepackage{latexsym}

\newcommand\blfootnote[1]{%
  \begingroup
  \renewcommand\thefootnote{}\footnote{#1}%
  \addtocounter{footnote}{-1}%
  \endgroup
}

\usepackage[T1]{fontenc}

\usepackage[utf8]{inputenc}

\usepackage{microtype}
\usepackage{booktabs,adjustbox,amsmath,amssymb}
\usepackage{xcolor,colortbl,multirow}
\usepackage{tikz,tikz-qtree}
\usepackage{enumitem}

\title{Low-Complexity Probing via Finding Subnetworks}

\author{Steven Cao$^{1,2}$ \qquad Victor Sanh$^{2}$ \qquad Alexander M. Rush$^{2}$ \\
	\\
	$^1$University of California, Berkeley \\
	$^2$Hugging Face \\
	{\tt stevencao@berkeley.edu} \\
	{\tt \{victor,sasha\}@huggingface.co}}

\date{}

\begin{document}
	\maketitle
	\begin{abstract}
		
		The dominant approach in probing neural networks for linguistic properties is to train a new shallow multi-layer perceptron (MLP) on top of the model's internal representations. This approach can detect properties encoded in the model, but at the cost of adding new parameters that may learn the task directly. We instead propose a subtractive pruning-based probe, where we find an existing subnetwork that performs the linguistic task of interest. Compared to an MLP, the subnetwork probe achieves both higher accuracy on pre-trained models and lower accuracy on random models, so it is both better at finding properties of interest and worse at learning on its own. Next, by varying the complexity of each probe, we show that subnetwork probing Pareto-dominates MLP probing in that it achieves higher accuracy given any budget of probe complexity. Finally, we analyze the resulting subnetworks across various tasks to locate where each task is encoded, and we find that lower-level tasks are captured in lower layers, reproducing similar findings in past work.
	\end{abstract}
	
	\section{Introduction}
	While pre-training has produced large gains for natural language tasks, it is unclear what a model learns during pre-training. Research in \textit{probing} investigates this question by training a shallow classifier on top of the pre-trained model's internal representations to predict some linguistic property~\citep[][\textit{inter alia}]{adi2016finegrained,shi-etal-2016-string,tenney-etal-2019-bert}. The resulting accuracy is then roughly indicative of the model encoding that property.\blfootnote{The code is available at \url{https://github.com/stevenxcao/subnetwork-probing}.}
	
	However, it is unclear how much is learned by the probe versus already captured in the model representations. This question has been the subject of much recent debate~\citep[][\textit{inter alia}]{hewitt-liang-2019-designing,voita2020informationtheoretic,pimentel-etal-2020-information}. We would like the probe to find only and all properties captured by a model, leading to a tradeoff between accuracy and complexity: a linear probe is insufficient to find the non-linear patterns in neural models, but a deeper multi-layer perceptron (MLP) is complex enough to learn the task on its own. 
	
	Motivated by this tradeoff and the goal of low-complexity probes, we consider a different approach based on pruning. Specifically, we search for a subnetwork --- a version of the model with a subset of the weights set to zero --- that performs the task of interest. As our search procedure, we build upon past work in pruning and perform gradient descent on a continuous relaxation of the search problem~\citep{louizos2017learning,mallya2018piggyback,sanh2020movement}. The resulting probe has many fewer free parameters than MLP probes.
	
	Our experiments evaluate the accuracy-complexity tradeoff compared to MLP probes on an array of linguistic tasks. First, we find that the neuron subnetwork probe has both higher accuracy on pre-trained models and lower accuracy on random models, so it is both better at finding properties of interest and less able to learn the tasks on its own. Next, we measure complexity as the bits needed to transmit the probe parameters~\citep{pimentel-etal-2020-pareto,voita2020informationtheoretic}. Varying the complexity of each probe, we find that subnetwork probing Pareto-dominates MLP probing in that it achieves higher accuracy given any desired complexity. Finally, we analyze the resulting subnetworks across various tasks and find that lower-level tasks are captured in lower layers, reproducing similar findings in past work~\citep{tenney-etal-2019-bert}. These results suggest that subnetwork probing is an effective new direction for improving our understanding of pre-training.

	\section{Related Work}
	
	\paragraph{Probing.}
	
	Probing investigates whether a model captures some hypothesized property and typically involves learning a shallow classifier on top of the model's frozen internal representations~\citep{adi2016finegrained,shi-etal-2016-string,conneau-etal-2018-cram}. Recent work has primarily applied this technique to pre-trained models.\footnote{While probing is also used in other domains (e.g.\ neural decoding), we focus on understanding neural models. Therefore, one source of strength for our probe is that we exploit the entire model, rather than only operating on representations.} \citet{clark-etal-2019-bert}, \citet{hewitt-manning-2019-structural}, and \citet{Manning2020EmergentLS} found that BERT captures various properties of syntax. \citet{tenney-etal-2019-bert} probed the layers of BERT for an array of tasks, and they found that their localization mirrored the classical NLP pipeline (part-of-speech, parsing, named entity recognition, semantic roles, coreference) in that lower-level tasks were captured in the lower layers.
	
	However, these results are difficult to interpret due to the use of a learned classifier. One line of work suggests comparing the probe accuracy to random baselines, e.g.\ random models~\citep{zhang-bowman-2018-language} or random control tasks~\citep{hewitt-liang-2019-designing}. Other works take an information-theoretic view: \citet{voita2020informationtheoretic} measure the complexity of the probe in terms of the bits needed to transmit its parameters, while \citet{pimentel-etal-2020-information} argue that probing should measure mutual information between the representation and the property. \citet{pimentel-etal-2020-pareto} propose a Pareto approach where they plot accuracy versus probe complexity, unifying several of these goals. We use these proposed metrics to compare our probing method to standard probing approaches.
	
	\paragraph{Subnetworks.}
	
	While pruning is widely used for model compression,
	some works have explored pruning as a technique for learning as well. \citet{mallya2018piggyback} found that a model trained on ImageNet could be used for new tasks by learning a binary mask over the weights. More recently, \citet{pmlr-v108-radiya-dixit20a} and \citet{zhao-etal-2020-masking} showed the analogous result in NLP that weight pruning can be used as an alternative to fine-tuning for pre-trained models. Our paper seeks to use pruning to reveal what the model already captures, rather than learn new tasks.
	
	\section{Subnetwork Probing}
	
	Given a task and a pre-trained encoder model with a classification head, our goal is to find a subnetwork with high accuracy on that task, where a subnetwork is the model with a subset of the encoder weights masked, i.e.\ set to zero. We search for this subnetwork via supervised gradient descent on the head and a continuous relaxation of the mask. We also mask at several levels of granularity, including pruning weights, neurons, or layers.
	
	To learn the masks, we follow \citet{louizos2017learning}. Letting $\phi \in \mathbb{R}^d$ denote the model weights, we associate the $i$th weight $\phi_i$ with a real-valued parameter $\theta_i$, which parameterizes a random variable $Z_i \in [0,1]$ representing the mask. $Z_i$ follows the hard concrete distribution $\text{HardConcrete}(\beta, \theta_i)$ with temperature $\beta$ and location $\theta_i$,
	\begin{align*}
	U_i &\sim \text{Unif}[0,1] \\
	S_i &= \sigma\left( \frac{1}{\beta} \left( \log \frac{U_i}{1-U_i} + \theta_i \right) \right) \\
	Z_i &= \min \left( 1, \max \left( 0, S_i(\zeta - \gamma) + \gamma \right) \right),
	\end{align*}
	where $\sigma$ denotes the sigmoid and $\gamma = -0.1$, $\zeta = 1.1$ are constants. This random variable can be thought of as a soft version of the Bernoulli. $S_i$ follows the concrete (or Gumbel-Softmax) distribution with temperature $\beta$~\citep{maddison2016concrete,jang2016categorical}. To put non-zero mass on $0$ and $1$, the distribution is stretched to the interval $(\gamma = -0.1, \zeta = 1.1)$ and clamped back to $[0,1]$.
	
	We will denote the mask as $Z_i = z(U_i, \theta_i)$ and the masked weights as $\phi * Z$. We can then optimize the mask parameters $\theta$ via gradient descent. Specifically, let $f(x; \phi)$ denote the model. Then, given a data point $(x,y)$ and a loss function $L$, we can minimize the expectation of the loss, or
	\begin{align*}
	\mathcal{L}(x,y,\theta) = \mathbb{E}_{U_i \sim \text{Unif}[0,1]} L(f(x; \phi * z(U, \theta)), y).
	\end{align*}
	We estimate the expectation via sampling: we sample a single $U$ and take the gradient $\nabla_\theta L(f(x; \phi * z(U, \theta))$. To encourage sparsity, we penalize the mask based on the probability it is non-zero, or
	\begin{align*}
	R(\theta) &= \mathbb{E} \|\theta\|_0 = \frac{1}{d} \sum_{i=1}^d \sigma\left( \theta_i - \beta \log \frac{-\gamma}{\zeta} \right).
	\end{align*}
	Letting $\lambda$ denote regularization strength, our objective becomes
	$\frac{1}{|D|}\sum_{(x,y) \in D} \mathcal{L}(x, y, \theta) + \lambda R(\theta)$.\footnote{ Departing from past work, we schedule $\lambda$ linearly to improve search: it stays fixed at $0$ for the first 25\% of training, linearly increases to $\lambda_\text{max}$ for the next 50\%, and then stays fixed. We set $\lambda_{\text{max}} = 1$ in our evaluation experiments.}

	\section{Probe Evaluation}
	
	To evaluate the accuracy-complexity tradeoff of a probe, we adapt methodology from recent work. First, we consider the non-parametric test of probing a random model~\citep{zhang-bowman-2018-language}. We check probe accuracy on the pre-trained model, the model with the encoder randomly reset (\textit{reset encoder}), and the model with the encoder and embeddings reset (\textit{reset all}). An ideal probe should achieve high accuracy on the pre-trained model and low accuracy on the reset models.\footnote{The reset encoder model contains some non-contextual information from its word embeddings, but no modeling of context; therefore, we would expect it to have better probe accuracy on tasks based mainly on word type (e.g.\ part-of-speech tagging).}
	
	Next, we consider a parametric test based on probe complexity. We first vary the 
	complexity of each probe, where for subnetwork probing we associate multiple encoder weights with a single mask,\footnote{For subnetworks, the pre-trained model has 72 matrices of size $768 \times 768$; see \url{https://github.com/huggingface/transformers/blob/v3.4.0/src/transformers/modeling_bert.py}. For each matrix, let $n_r$ and $n_c$ denote the number of rows and columns per mask. Then, we set $(n_r, n_c)$ to $(768,768)$, $(768,192)$, $(768,24)$, $(768,6)$, $(768,1)$, $(192,1)$, $(24,1)$, $(6,1)$, and $(1,1)$. $(768, 768)$ corresponds to masking entire matrices, $(768, 1)$ to masking neurons, and $(1,1)$ to masking weights.} and for the MLP probe we restrict the rank of the hidden layer. We then plot the resulting accuracy-complexity curve~\citep{pimentel-etal-2020-pareto}. 
	
	To plot this curve, we need a measure of complexity that can compare probes of different types. Therefore, we measure complexity as the number of bits needed to transmit the probe parameters~\citep{voita2020informationtheoretic}, where for simplicity we use a uniform encoding. In the subnetwork case, this encoding corresponds to using a single bit for each mask parameter. In the case of an MLP probe, each parameter is a real number, so the number of bits per parameter depends on its range and precision. For example, if each parameter lies in $[a,b]$ and requires $\epsilon$ precision, then we need $\log (\frac{b-a}{\epsilon})$ bits per parameter. To avoid having the choice of precision impact results, we plot lower and upper bounds of $1$ and $32$ bits per parameter.

	\section{Experimental Setup}
	
	We probe \texttt{bert-base-uncased}~\citep{devlin_bert:2018,wolf-etal-2020-transformers} for the following tasks:
	
	(1) \textbf{Part-of-speech Tagging:} We use the part-of-speech tags in the universal dependencies dataset~\citep{zeman-etal-2017-conll}. As our classification head, we use dropout with probability $p = 0.1$, followed by a linear layer and softmax projecting from the BERT dimension to the number of tags.
	
	(2) \textbf{Dependency Parsing}: We use the universal dependencies dataset~\citep{zeman-etal-2017-conll} and the biaffine head for classification~\citep{dozat2016deep}. We report macro-averaged labeled attachment score.
	
	(3) \textbf{Named Entity Recognition (NER)}: We use the data from the CoNLL 2003 shared task~\citep{tjong-kim-sang-de-meulder-2003-introduction} and the same classification head as for part-of-speech tagging. We report F1 using the CoNLL 2003 script.
	
	Our primary probing baseline is the MLP probe with one hidden layer (MLP-1):
	\begin{align*}
	\texttt{MLP-1}(x) &= \texttt{ReLU}(\texttt{LayerNorm}(UV^Tx)),
	\end{align*}
	with $U, V \in \mathbb{R}^{d \times r}$.
	The choice of $r$ restricts the rank of the hidden layer and thus its complexity.\footnote{We set the rank to $1$, $2$, $5$, $10$, $25$, $50$, $125$, $250$, and $768$.} Then, if $g(x; \phi)$ is our pre-trained encoder and $\texttt{cls}$ is the classification head, our two probes are $f_{\text{Subnetwork}}(x) = \texttt{cls}( g(x; \phi * Z))$ and {$f_{\text{MLP-1}}(x) = \texttt{cls}(\texttt{MLP-1}(g(x; \phi)))$}. 
	
	While we vary the complexity of each probe to produce the accuracy-complexity plot, we default to neuron subnetwork probing and full rank MLP-1 probing in all other experiments.
	
	\begin{table}[t]
		\begin{center}
			\resizebox{\linewidth}{!}{\begin{tabular}{lccc}
					\toprule
					& Pre-trained $\uparrow$  & Reset encoder $\downarrow$& Reset all $\downarrow$ \\
					\midrule
					\multicolumn{4}{l}{Part-of-speech Tagging} \\
					\midrule
					Subnetworks & \textbf{93.39} & 87.53 & 71.53 \\
					MLP-1 & 90.25 & 86.53 & \textbf{69.16} \\
					\midrule
					Fine-tuning &  95.69 & 86.47 & 84.42 \\
					\midrule
					\midrule
					\multicolumn{4}{l}{Dependency Parsing} \\
					\midrule
					Subnetworks & \textbf{86.86} & 54.31 & \textbf{39.84} \\
					MLP-1 & 76.65 & 54.09 & 42.81  \\
					\midrule
					Fine-tuning & 89.93 & 79.10 & 74.48 \\
					\midrule
					\midrule
					\multicolumn{4}{l}{Named Entity Recognition} \\
					\midrule
					Subnetworks & \textbf{87.94} & \textbf{68.09} & \textbf{30.83} \\
					MLP-1 & 84.80 & 69.35 & 53.25  \\
					\midrule
					Fine-tuning &  93.68 & 81.80  & 70.08 \\
					\bottomrule
			\end{tabular}}
		\end{center}
		\caption{\label{table:probing} Probe accuracy for \texttt{bert-base-uncased} (Pre-trained), the model with the encoder reset but the embeddings preserved (Reset encoder), and the model completely reset (Reset all). The $\uparrow$ and $\downarrow$ denote whether higher or lower is better (substantially better numbers are bolded). For reference, we also include fine-tuning (training all model parameters rather than probing). Compared to MLP-1, neuron subnetwork probing achieves higher accuracy for the pre-trained model and lower accuracy for the random models.}
	\end{table}
	
	\begin{figure*}[t]
		\begin{center}
			\includegraphics[width=\linewidth]{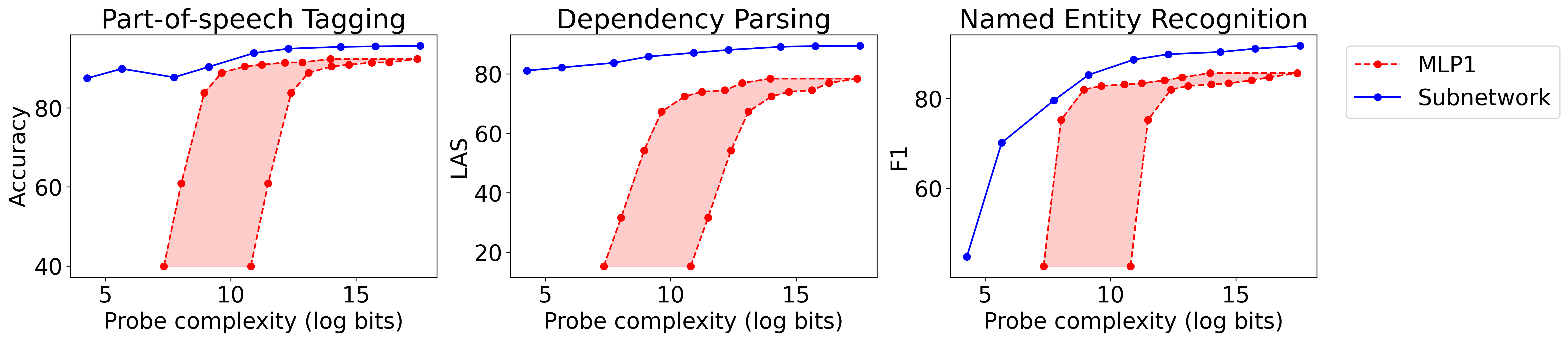}
		\end{center}
		\caption{\label{figure:pareto} Subnetwork probe and MLP-1 probe accuracy on the pre-trained model plotted versus probe complexity, measured in $\ln(\text{bits})$. For the MLP-1 probe, we plot lower and upper bounds on complexity of $1$ and $32$ bits per parameter. The subnetwork probe Pareto-dominates the MLP-1 probe in that it achieves higher accuracy for any desired complexity, even if we assume the optimistic lower bound on MLP-1 complexity of $1$ bit per parameter. }
	\end{figure*}
	
	\begin{figure*}[t]
		\begin{center}
			\includegraphics[width=\linewidth]{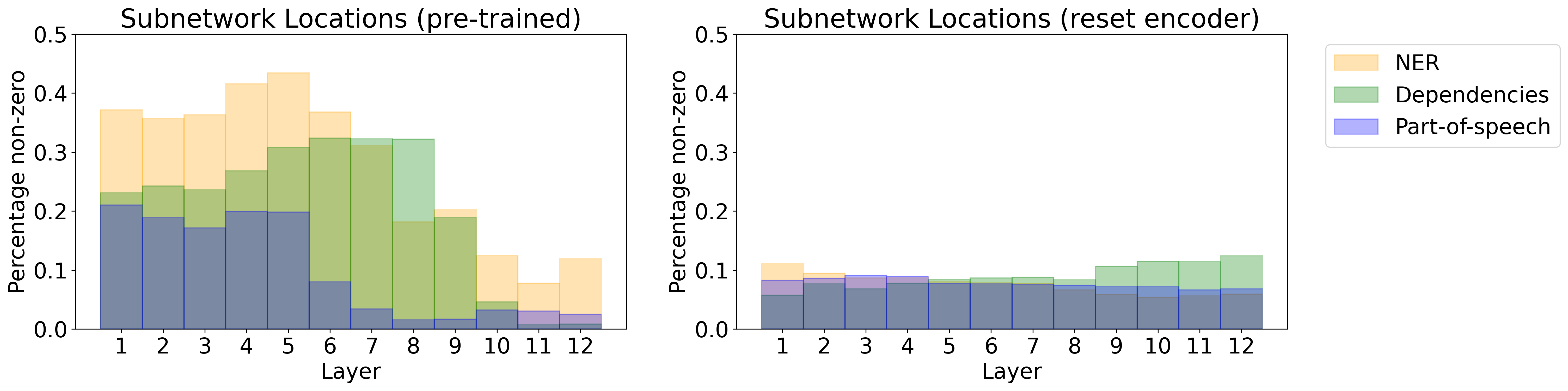}
		\end{center}
		\caption{\label{figure:location} The percentage of non-zero weights in each layer for subnetworks of the pre-trained model and the reset encoder model. While the reset encoder model's subnetworks are distributed uniformly across the layers, the pre-trained model's subnetworks are localized, with the order part-of-speech $\to$ dependencies $\to$ NER.}
	\end{figure*}
	
	\section{Results}
	
	\paragraph{Accuracy-Complexity Tradeoff.} 
	
	Table~\ref{table:probing} shows the results from the non-parametric experiments. When probing the pre-trained model, the subnetwork probe has much higher accuracy than the MLP-1 probe across all tasks. Furthermore, when probing the random models, the subnetwork probe has much lower accuracy for dependency parsing and NER, suggesting that the probe is less able to learn the task on its own. Overall, these numbers suggest that the subnetwork probe is a more faithful probe in that it finds properties when they are present, and does not find them in a random model.
	
	Figure~\ref{figure:pareto} plots the results from the parametric experiments, where we vary the complexity of each probe, apply it to the pre-trained model, and plot the resulting accuracy-complexity curve. We find that the subnetwork probe Pareto-dominates the MLP-1 probe in that it achieves higher accuracy for any complexity, even if we assume an overly optimistic MLP-1 lower bound of 1 bit per parameter. In particular, for part-of-speech and dependency parsing, the subnetwork probe achieves high accuracy even when given only 72 bits, while the MLP-1 probe falls off heavily at $\sim$20K bits.
	
	\paragraph{Subnetwork Analysis.}
	
	An auxiliary benefit of subnetwork probing is that we can examine the subnetworks produced by the procedure. One possibility is to look at the locations of the subnetworks, and one way to examine location is to count the number of unmasked weights in each layer. Figure~\ref{figure:location} shows locations of the remaining parameters
	in the subnetworks extracted from the pre-trained model and the random encoder model. To prune as many parameters as possible, we set $\lambda_\text{max}$ to be the largest out of $(1, 5, 25, 125)$ such that accuracy is within 10\% of fine-tuning accuracy (see the Appendix for more details). We then examine the sparsity levels of the attention heads for each layer.
	While reset encoder model's subnetworks are uniformly distributed across the layers, the pre-trained model's subnetworks are localized and follow the order part-of-speech $\to$ dependencies $\to$ NER, reproducing the order found in \citet{tenney-etal-2019-bert}. While \citet{tenney-etal-2019-bert} derived layer importance by training classifiers at each layer, we find location directly via pruning. This experiment strengthens their result and represents one example where subnetwork probing reveals additional insights into the model beyond accuracy.
	
	\section{Conclusion}
	Together, these results show that subnetwork probing is more faithful to the model and offers richer analysis than existing probing approaches. While this work explores accuracy and location-based analysis, there are other possible directions, e.g., applying neuron explainability techniques. Therefore, we see subnetwork probing as a fruitful new direction for understanding pre-training.
	
	\section{Ethical Considerations}
	While pre-trained models have improved performance for many NLP tasks, they exhibit biases present in the pre-training corpora~\citep[][\textit{inter alia}]{manzini-etal-2019-black,tan2019assessing,kurita-etal-2019-measuring}. As a result, deploying pre-trained models runs the risk of reinforcing social biases. Probing gives us a tool to better understand and hopefully mitigate these biases. As one example of such a study, \citet{vig2020investigating} analyze how neurons and attention heads contribute to gender bias in pre-trained transformers. Therefore, while we analyze linguistic tasks in our paper, our method could also provide insights into model bias, e.g.\ by analyzing subnetworks for bias detection tasks like CrowS-Pairs~\citep{nangia-etal-2020-crows} or StereoSet~\citep{nadeem2020stereoset}.
	
	\section{Acknowledgements}
	We would like to thank Eric Wallace, Kevin Yang, Ruiqi Zhong, Dan Klein, and Yacine Jernite for their useful comments and feedback. This work was done during an internship at Hugging Face.
	
	\bibliographystyle{acl_natbib}
	\bibliography{custom}
	
	\appendix
	\section{Appendix}
	\subsection{Hyperparameters}
	The mask parameters are optimized using Adam with $\beta = (0.9, 0.999)$, $\epsilon = 1\times 10^{-8}$, and learning rate $0.2$ with linear warmup for the first $10\%$ of the data. The classification head parameters are also optimized using Adam with the same hyperparameters and warmup, except with learning rate $5 \times 10^{-5}$. The MLP-1 and fine-tuning baselines are also optimized using Adam with the same hyperparameters, warmup, and learning rate $5 \times 10^{-5}$.  We train for 30 epochs for all tasks.
	
	\subsection{Varying Regularization Strength}
	
	\begin{table}[t]
		\begin{center} 
			\resizebox{\linewidth}{!}{\begin{tabular}{lccc}
					\toprule
					& Reset all & Reset encoder & Pre-trained \\
					\midrule
					\midrule
					\multicolumn{4}{l}{Part-of-speech Tagging} \\
					\midrule
					Subnetworks \\
					$\lambda_\text{max}=1$& 71.53 & 87.53 & 93.39 \\
					$\lambda_\text{max}=5$& 70.45 & 87.20 & 92.41 \\
					$\lambda_\text{max}=25$& 68.65 & 86.23 & 90.66 \\
					$\lambda_\text{max}=125$& 66.39 & 86.10 & 84.86\\
					\midrule
					MLP-1 & 69.16 & 86.53 & 90.25 \\
					Fine-tuning & 84.42 & 86.47 & 95.69 \\
					\midrule
					\midrule
					\multicolumn{4}{l}{Dependency Parsing} \\
					\midrule
					Subnetworks \\
					$\lambda_\text{max}=1$ & 39.84  & 54.31  & 86.86 \\
					$\lambda_\text{max}=5$ & 43.36  & 54.93  & 85.99 \\
					$\lambda_\text{max}=25$& 41.43  & 54.41  & 83.12\\
					$\lambda_\text{max}=125$& 43.07 & 53.70 & 74.49\\
					\midrule
					MLP-1 & 42.81  & 54.09  & 76.65 \\
					Fine-tuning & 74.48  & 79.10 & 89.93 \\
					\midrule
					\midrule
					\multicolumn{4}{l}{Named Entity Recognition (NER)} \\
					\midrule
					Subnetworks \\
					$\lambda_\text{max} = 1$ & 30.83  & 68.09  & 87.94 \\
					$\lambda_\text{max} = 5$ & 22.92  & 65.18  & 84.48 \\
					$\lambda_\text{max}=25$& \phantom{0}3.41  & 57.82  & 72.26 \\	
					$\lambda_\text{max}=125$& \phantom{0}2.04 & 57.56 & 50.67 \\
					\midrule				
					MLP-1 & 53.25  & 69.35  & 84.80 \\
					Fine-tuning & 70.08  & 81.80  & 93.68 \\
					\bottomrule
			\end{tabular}}
		\end{center}
		\caption{\label{table:varyinglambda} Subnetwork probing accuracies while varying regularization strength. }
	\end{table}
	
	Table~\ref{table:varyinglambda} shows probing accuracies for $\lambda_\text{max} \in (1, 5, 25, 125)$. Our method is consistently more selective than MLP-1 across the various values of $\lambda_\text{max}$, except for $\lambda_\text{max} = 125$, which seems to require too much sparsity. 
	
	\subsection{Reproducibility Checklist}
	
	Experiments were run in Google Colab using a single 12GB NVIDIA Tesla K80 GPU. For each task, one run of fine-tuning took about half an hour. We used the transformers implementation of the \texttt{bert-base-uncased} model~\citep{wolf-etal-2020-transformers,devlin_bert:2018}, which has 12 layers, 768 hidden dimension, 12 heads, and 110M parameters. As data, we used the dev (2002 examples) and train (12541 examples) splits of the English universal dependencies dataset~\citep{zeman-etal-2017-conll}, and the test (3235 examples) and train (13862 examples) splits of the CoNLL 2003 NER shared task~\citep{tjong-kim-sang-de-meulder-2003-introduction}.
	
\end{document}